\documentclass[11pt,a4paper]{article}
\usepackage[hyperref]{emnlp-ijcnlp-2019}
\usepackage{times}
\usepackage{latexsym}
\usepackage{wasysym}
\usepackage{comment}
\usepackage{url}
\usepackage{float}
\usepackage{booktabs}
\usepackage{graphicx}
\usepackage{pgfplots}
\usepackage{tikz-dependency}
\usepackage{multirow}
\pgfplotsset{width=0.47\textwidth,compat=1.14, height=2in}

\definecolor{PQgreen}{rgb}{0.3,0.8,0.44}

\definecolor{red}{HTML}{E31A1C}
\definecolor{blue}{HTML}{1F78B4}
\definecolor{green}{HTML}{33A02C}
\definecolor{orange}{HTML}{FF7F00}
\definecolor{purple}{HTML}{6A3D9A}

\aclfinalcopy%

\newcommand\golden{\textsc{GoldEn}}

\newcommand\eg{{\it e.g.}}
\newcommand{\fone}{F\textsubscript{1}}

\newcommand\hotpotqa{\textsc{HotpotQA}}
\newcommand\qangaroo{QAngaroo}
\newcommand\squad{SQuAD}
\newcommand\triviaqa{TriviaQA}
\newcommand{\fake}[1]{{\color{red} #1}}
\renewcommand{\fake}[1]{#1}
\usepackage{soul}

\title{Answering Complex Open-domain Questions \\Through Iterative Query Generation}

\author{
  Peng Qi$^\dagger$\quad Xiaowen Lin$^{*\dagger}$ \quad Leo Mehr$^{*\dagger}$ \quad Zijian Wang$^{*\ddagger}$ \quad Christopher D. Manning$^\dagger$\\
  $\dagger$ Computer Science Department \quad $\ddagger$ Symbolic Systems Program\\
  Stanford University\\
  {\tt \{pengqi, veralin, leomehr, zijwang, manning\}@cs.stanford.edu} \\}

\date{}

\begin{document}

\setlength{\abovedisplayskip}{3pt}
\setlength{\belowdisplayskip}{3pt}

\maketitle
\begin{abstract}
  It is challenging for current one-step retrieve-and-read question answering (QA) systems to answer questions like \emph{``Which novel by the author of `Armada' will be adapted as a feature film by Steven Spielberg?''}\ because the question seldom contains retrievable clues about the missing entity (here, the author).
Answering such a question requires multi-hop reasoning where one must gather information about the missing entity (or facts) to proceed with further reasoning.
We present \golden{} (Gold Entity) Retriever, which iterates between reading context and retrieving more supporting documents to answer open-domain multi-hop questions.
Instead of using opaque and computationally expensive neural retrieval models, \golden{} Retriever generates natural language search queries given the question and available context, and leverages off-the-shelf information retrieval systems to query for missing entities.
This allows \golden{} Retriever to scale up efficiently for open-domain multi-hop reasoning while maintaining interpretability.
We evaluate \golden{} Retriever on the recently proposed open-domain multi-hop QA dataset, \hotpotqa{}, and demonstrate that it outperforms the best previously published model despite not using pretrained language models such as BERT.

\end{abstract}

\renewcommand{\thefootnote}{\fnsymbol{footnote}}
\footnotetext[1]{Equal contribution, order decided by a random number generator.}
\renewcommand{\thefootnote}{\arabic{footnote}}

\section{Introduction}

\begin{figure}
    \centering
    \includegraphics[width=0.48\textwidth]{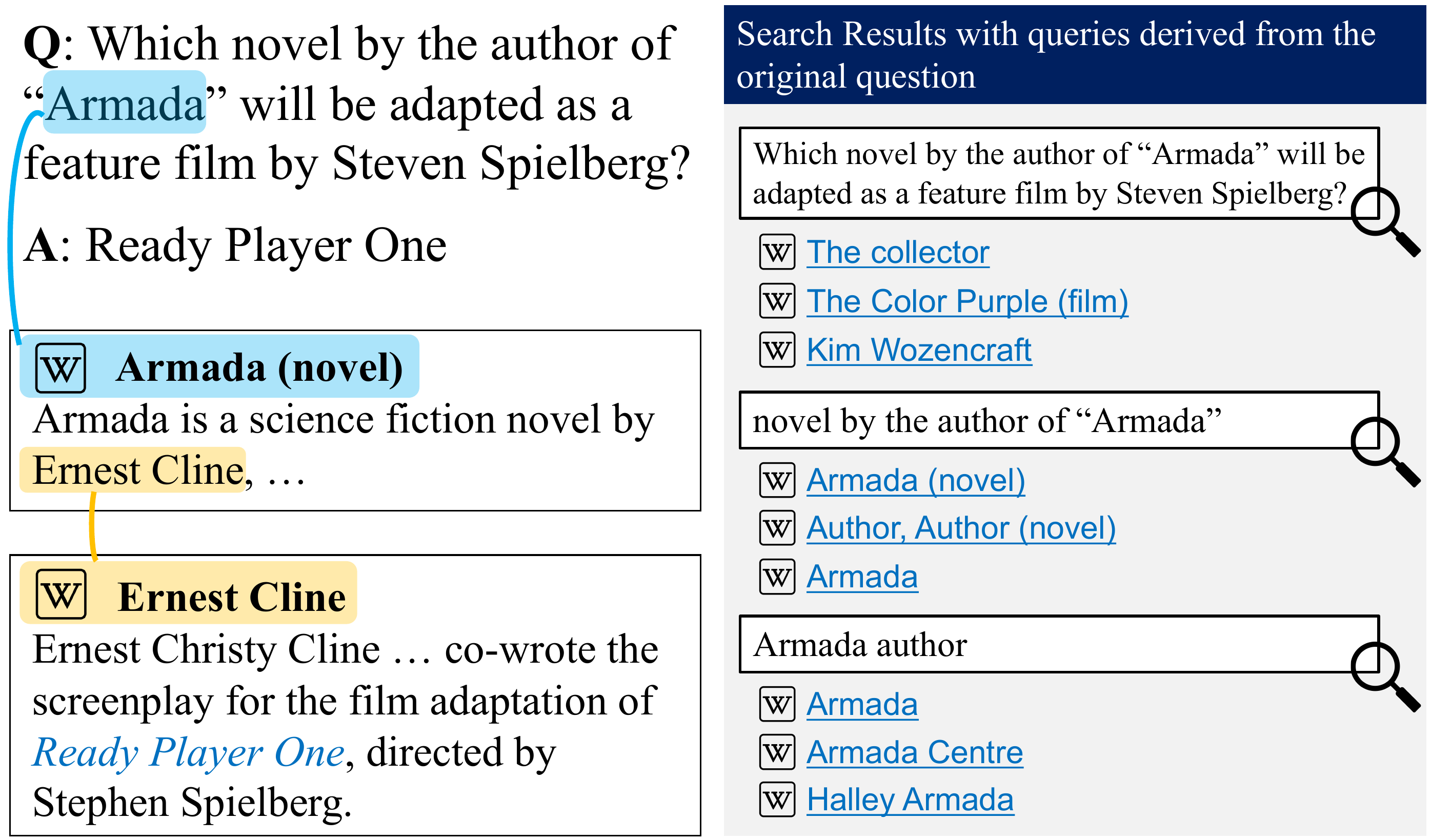}
    \caption{An example of an open-domain multi-hop question from the \hotpotqa{} dev set, where ``Ernest Cline'' is the missing entity. Note from the search results that it cannot be easily retrieved based on merely the question. (Best viewed in color)} \label{fig:example}
\end{figure}

Open-domain question answering (QA) is an important means for us to make use of knowledge in large text corpora and enables diverse queries without requiring a knowledge schema ahead of time.
Enabling such systems to perform multi-step inference%
can further expand our capability to explore the knowledge in these corpora (\eg, see Figure \ref{fig:example}).

Fueled by the recently proposed large-scale QA datasets such as \squad{} \cite{rajpurkar2016squad, rajpurkar2018know} and \triviaqa{} \cite{JoshiTriviaQA2017}, much progress has been made in open-domain question answering.
\citet{chen2017reading} proposed a two-stage approach of retrieving relevant content with the question, then reading the paragraphs returned by the information retrieval (IR) component to arrive at the final answer.
This ``\emph{retrieve and read}'' approach has since been adopted and extended in various open-domain QA systems \cite{nishida2018retrieve, Kratzwald2018AdaptiveDR}, but it is inherently limited to answering questions that do not require multi-hop/multi-step reasoning.
This is because for many multi-hop questions, not all the relevant context can be obtained in a single retrieval step (\eg, ``Ernest Cline'' in Figure \ref{fig:example}).

More recently, the emergence of multi-hop question answering datasets such as \qangaroo{} \cite{welbl2018constructing} and \hotpotqa{} \cite{yang2018hotpotqa} has sparked interest in multi-hop QA in the research community.
Designed to be more challenging than \squad{}-like datasets, they feature questions that require context of more than one document to answer, testing QA systems' abilities to infer the answer in the presence of multiple pieces of evidence and to efficiently find the evidence in a large pool of candidate documents.
However, since these datasets are still relatively new, most of the existing research focuses on the few-document setting where a relatively small set of context documents is given, which is guaranteed to contain the ``gold'' context documents, all those from which the answer comes \cite{decao2019question, zhong2019coarse}.

In this paper, we present \golden{} (Gold Entity) Retriever.
Rather than relying purely on the original question to retrieve passages, the central innovation is that at each step the model also uses IR results from previous hops of reasoning to generate a new natural language query and retrieve new evidence to answer the original question.
For the example in Figure \ref{fig:example}, \golden{} would first generate a query to retrieve \emph{Armada (novel)} based on the question, then query for \emph{Ernest Cline} based on newly gained knowledge in that article.
This allows \golden{} to leverage off-the-shelf, general-purpose IR systems to scale open-domain multi-hop reasoning to millions of documents efficiently, and to do so in an interpretable manner.
Combined with a QA module that extends BiDAF++ \cite{clark2017simple}, our final system outperforms the best previously published system on the open-domain (fullwiki) setting of \hotpotqa{} without using powerful pretrained language models like BERT \cite{devlin2018bert}.

The main contributions of this paper are: (a) a novel iterative retrieve-and-read framework capable of multi-hop reasoning in open-domain QA%
\footnote{Code and pretrained models available at \url{https://github.com/qipeng/golden-retriever}}
(b) a natural language query generation approach that guarantees interpretability in the multi-hop evidence gathering process;
(c) an efficient training procedure to enable query generation with minimal supervision signal that significantly boosts recall of gold supporting documents in retrieval.

\section{Related Work}

\paragraph{Open-domain question answering (QA)}
Inspired by the series of TREC QA competitions,\footnote{\url{http://trec.nist.gov/data/qamain.html}} \citet{chen2017reading} were among the first to adapt neural QA models to the open-domain setting.
They built a simple inverted index lookup with TF-IDF on the English Wikipedia, and used the question as the query to retrieve top 5 results for a reader model to produce answers with.
Recent work on open-domain question answering largely follow this retrieve-and-read approach, and focus on improving the information retrieval component with question answering performance in consideration \cite{nishida2018retrieve, Kratzwald2018AdaptiveDR, nogueira2019document}.
However, these one-step retrieve-and-read approaches are fundamentally ill-equipped to address questions that require multi-hop reasoning, especially when necessary evidence is not readily retrievable with the question.

\paragraph{Multi-hop QA datasets}
\qangaroo{} \cite{welbl2018constructing} and \hotpotqa{} \cite{yang2018hotpotqa} are among the largest-scale multi-hop QA datasets to date.
While the former is constructed around a knowledge base and the knowledge schema therein, the latter adopts a free-form question generation process in crowdsourcing and span-based evaluation.
Both datasets feature a few-document setting where the gold supporting facts are provided along with a small set of distractors to ease the computational burden.
However, researchers have shown that this sometimes results in gameable contexts, and thus does not always test the model's capability of multi-hop reasoning \cite{chen2019understanding, min2019compositional}.
Therefore, in this work, we focus on the fullwiki setting of \hotpotqa{}, which features a truly open-domain setting with more diverse questions.

\paragraph{Multi-hop QA systems}
At a broader level, the need for multi-step searches, query task decomposition, and subtask extraction has been clearly recognized in the IR community \cite{hassan2014supporting, mehrotra2016deconstructing, mehrotra2017extracting}, but multi-hop QA has only recently been studied closely with the release of large-scale datasets.
Much research has focused on enabling multi-hop reasoning in question answering models in the few-document setting, \eg, by modeling entity graphs \cite{decao2019question} or scoring answer candidates against the context \cite{zhong2019coarse}.
These approaches, however, suffer from scalability issues when the number of supporting documents and/or answer candidates grow beyond a few dozen.
\citet{ding2019cognitive} apply entity graph modeling to \hotpotqa{}, where they expand a small entity graph starting from the question to arrive at the context for the QA model.
However, centered around entity names, this model risks missing purely descriptive clues in the question.
\citet{das2019multistep} propose a neural retriever trained with distant supervision to bias towards paragraphs containing answers to the given questions, which is then used in a multi-step reader-reasoner framework.
This does not fundamentally address the discoverability issue in open-domain multi-hop QA, however, because usually not all the evidence can be directly retrieved with the question.
Besides, the neural retrieval model lacks explainability, which is crucial in real-world applications.
\citet{talmor2018web}, instead, propose to answer multi-hop questions at scale by decomposing the question into sub-questions and perform iterative retrieval and question answering, which shares very similar motivations as our work. However, the questions studied in that work are based on logical forms of a fixed schema, which yields additional supervision for question decomposition but limits the diversity of questions. More recently, \citet{min2019multi} apply a similar idea to \hotpotqa{}, but this approach similarly requires additional annotations for decomposition, and the authors did not apply it to iterative retrieval.

\begin{figure*}[!ht]
    \centering
    \includegraphics[width=0.98\textwidth]{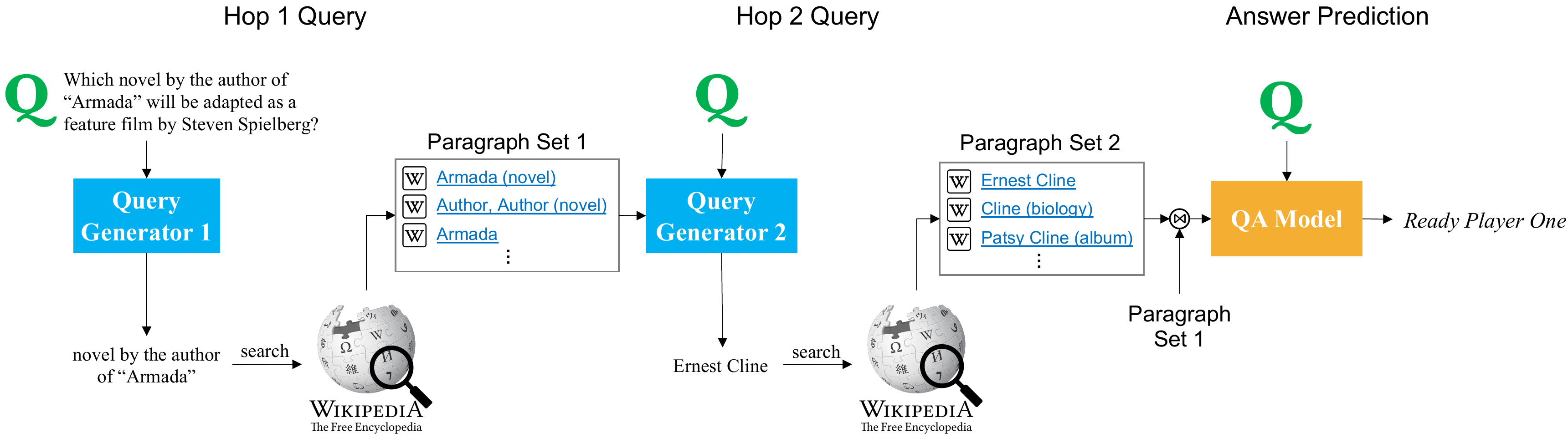}
    \caption{Model overview of \golden{} Retriever.
    Given an open-domain multi-hop question, the model iteratively retrieves more context documents, and concatenates all retrieved context for a QA model to answer from.}
    \label{fig:model}
\end{figure*}

\section{Model}

In this section, we formally define the problem of open-domain multi-hop question answering, and motivate the architecture of the proposed \golden{} (Gold Entity) Retriever model.
We then detail the query generation components as well as how to derive supervision signal for them, before concluding with the QA component.

\subsection{Problem Statement}

We define the problem of \emph{open-domain multi-hop QA} as one involving a question $q$, and $S$ relevant (gold) supporting context documents $d_1, \ldots, d_S$ which contain the desired answer $a$%
\footnote{In this work, we only consider extractive, or span-based, QA tasks, but the problem statement and the proposed method apply to generative QA tasks as well.}
These supporting documents form a chain of reasoning necessary to arrive at the answer, and they come from a large corpus of documents $\mathcal{D}$ where $|\mathcal{D}| \gg S$.
In this chain of reasoning, the supporting documents are usually connected via shared entities or textual similarities (\eg, they describe similar entities or events), but these connections do not necessarily conform to any predefined knowledge schema.

We contrast this to what we call the \emph{few-document setting} of multi-hop QA, where the QA system is presented with a small set of documents $\mathcal{D}{\textsubscript{few-doc}}=\{d_1, \ldots, d_S, d'_1, \ldots, d'_D\}$, where $d'_1, \ldots, d'_D$ comprise a small set of ``distractor'' documents that test whether the system is able to pick out the correct set of supporting documents in the presence of noise.
This setting is suitable for testing QA systems' ability to perform multi-hop reasoning given the gold supporting documents with bounded computational budget, but we argue that it is far from a realistic one.
In practice, an open-domain QA system has to locate all gold supporting documents from $\mathcal{D}$ on its own, and as shown in Figure \ref{fig:example}, this is often difficult for multi-hop  questions based on the original question alone, as not all gold documents are easily retrievable given the question.

To address this gold context discoverability issue, we argue that it is necessary to move away from a single-hop retrieve-and-read approach where the original question serves as the search query.
In the next section, we introduce \golden{} Retriever, which addresses this problem by iterating between retrieving more documents and reading the context for multiple rounds.

\subsection{Model Overview}

Essentially, the challenge of open-domain multi-hop QA lies in the fact that the information need of the user $(q \to a)$ cannot be readily satisfied by any information retrieval (IR) system that models merely the similarity between the question $q$ and the documents.
This is because the true information need will only unfold with progressive reasoning and discovery of supporting facts.
Therefore, one cannot rely solely on a similarity-based IR system for such iterative reasoning, because the potential pool of relevant documents grows exponentially with the number of hops of reasoning.

To this end, we propose \golden{} (Gold Entity) Retriever, which makes use of the gold document\footnote{In \hotpotqa{}, documents usually describe entities, thus we use ``documents'' and ``entities'' interchangeably.} information available in the QA dataset at training time to iteratively query for more relevant supporting documents during each hop of reasoning.
Instead of relying on the original question as the search query to retrieve all supporting facts, or building computationally expensive search engines that are less interpretable to humans, we propose to leverage text-based IR engines for interpretability, and generate different search queries as each reasoning step unfolds.
In the very first hop of reasoning, \golden{} Retriever is presented the original question $q$, from which it generates a search query $q_1$ that retrieves supporting document $d_1$%
\footnote{For notational simplicity, $d_k$ denotes the supporting document needed to complete the $k$-th step of reasoning. We also assume that the goal of each IR query is to retrieve one and only one gold supporting document in its top $n$ results.}
Then for each of the subsequent reasoning steps ($k=2, \ldots, S$), \golden{} Retriever generates a query $q_k$ from the question and the available context, $(q, d_1, \ldots, d_{k-1})$.
This formulation allows the model to generate queries based on information revealed in the supporting facts (see Figure \ref{fig:model}, for example).

We note that \golden{} Retriever is much more efficient, scalable, and interpretable at retrieving gold documents compared to its neural retrieval counterparts.
This is because \golden{} Retriever does not rely on a QA-specific IR engine tuned to a specific dataset, where adding new documents or question types into the index can be extremely inefficient.
Further, \golden{} Retriever generates queries in natural language, making it friendly to human interpretation and verification.
One core challenge in \golden{} Retriever, however, is to train query generation models in an efficient manner, because the search space for potential queries is enormous and off-the-shelf IR engines are not end-to-end differentiable.
We outline our solution to this challenge in the following sections.

\subsection{Query Generation}

\begin{table*}
    \centering
    \resizebox{1\linewidth}{!}{
    \begin{tabular}{p{10.5cm}p{3.3cm}p{2.9cm}}
        \toprule
        \multicolumn{1}{c}{\textbf{Question}} & \textbf{Hop 1 Oracle} & \textbf{Hop 2 Oracle} \\
        \midrule
        What government position was held by the woman who portrayed Corliss Archer in the film Kiss and Tell? & Corliss Archer in the film Kiss and Tell & Shirley Temple \\
        \midrule
        Scott Parkin has been a vocal critic of Exxonmobil and another corporation that has operations in how many countries? & Scott Parkin & Halliburton \\
        \midrule
        Are Giuseppe Verdi and Ambroise Thomas both Opera composers? & Giuseppe Verdi & Ambroise Thomas\\
        \bottomrule
    \end{tabular}
    }
    \caption{Example oracle queries on the \hotpotqa{} dev set.}
    \label{tab:oracle_examples}
\end{table*}

For each reasoning step, we need to generate the search query given the original question $q$ and some context of documents we have already retrieved (initially empty).
This query generation problem is conceptually similar to the QA task in that they both map a question and some context to a target, only instead of an answer, the target here is a search query that helps retrieve the desired supporting document for the next reasoning step.
Therefore, we formulate the query generation process as a question answering task.

To reduce the potentially large space of possible queries, we favor a QA model that extracts text spans from the context over one that generates free-form text as search queries.
We therefore employ DrQA's Document Reader model \cite{chen2017reading}, which is a relatively light-weight recurrent neural network QA model that has demonstrated success in few-document QA. We adapt it to query generation as follows.

For each reasoning step $k = 1, \ldots, S$, given a question $q$ and some \emph{retrieval context} $C_k$ which \emph{ideally} contains the gold supporting documents $d_1, \ldots, d_{k-1}$, we aim to generate a search query $q_k$ that helps us retrieve $d_k$ for the next reasoning step.
A Document Reader model is trained to select a span from $C_k$ as the query
\[q_k = G_k(q, C_k),\]
where $G_k$ is the query generator. This query is then used to search for supporting documents, which are concatenated with the current retrieval context to update it
$$C_{k+1} = C_k \bowtie \textrm{IR}_n(q_k) $$
where $\textrm{IR}_n(q_k)$ is the top $n$ documents retrieved from the search engine using $q_k$, and $C_1=q$%
\footnote{In the query result, the title of each document is delimited with special tokens \texttt{<t>} and \texttt{</t>} before concatenation.}
At the end of the retrieval steps, we provide $q$ as question along with $C_S$ as context to the final few-document QA component detailed in Section \ref{sec:qa-component} to obtain the final answer to the original question.

To train the query generators, we follow the steps above to construct the retrieval contexts, but during training time, when $d_k$ is not part of the IR result, we replace the lowest ranking document with $d_k$ before concatenating it with $C_k$ to make sure the downstream models have access to necessary context.

\subsection{Deriving Supervision Signal for Query Generation}
When deriving supervision signal to train our query generators, the potential search space is enormous for each step of reasoning even if we constrain ourselves to predicting spans from the context.
This is aggravated by multiple hops of reasoning required by the question.
One solution to this issue is to train the query generators with reinforcement learning (RL) techniques (\eg, REINFORCE \cite{sutton2000policy}), where \cite{nogueira2017task} and \cite{buck2017ask} are examples of one-step query generation with RL. However, it is computationally inefficient, and has high variance especially for the second reasoning step and forward, because the context depends greatly on what queries have been chosen previously and their search results.

Instead, we propose to leverage the limited supervision we have about the gold supporting documents $d_1, \ldots, d_S$ to narrow down the search space.
The key insight we base our approach on is that at any step of open-domain multi-hop reasoning, there is some \emph{semantic overlap} between the retrieval context and the next document(s) we wish to retrieve.
For instance, in our \emph{Armada} example, when the retrieval context contains only the question, this overlap is the novel itself; after we have expanded the retrieval context, this overlap becomes the name of the author, \emph{Ernest Cline}.
Finding this semantic overlap between the retrieval context and the desired documents not only reveals the chain of reasoning naturally, but also allows us to use it as the search query for retrieval.

Because off-the-shelf IR systems generally optimize for shallow lexical similarity between query and candidate documents in favor of efficiency,
a good proxy for this overlap is locating spans of text that have high lexical overlap with the intended supporting documents.
To this end, we propose a simple yet effective solution, employing several heuristics to generate candidate queries: computing the longest common string/sequence between the current retrieval context and the title/text of the intended paragraph ignoring stop words, then taking the contiguous span of text that corresponds to this overlap in the retrieval context.
This allows us to not only make use of entity names, but also textual descriptions that better lead to the gold entities.
It is also more generally applicable than question decomposition approaches \cite{talmor2018web,min2019multi}, and does not require additional annotation for decomposition.

Applying various heuristics results in a handful of candidate queries for each document, and we use our IR engine (detailed next) to rank them based on recall of the intended supporting document to choose one as the final oracle query we train our query generators to predict.
This allows us to train the query generators in a fully supervised manner efficiently.
Some examples of oracle queries on the \hotpotqa{} dev set can be found in Table~\ref{tab:oracle_examples}.
We refer the reader to Appendix \ref{sec:heuristics} for more technical details about our heuristics and how the oracle queries are derived.

\paragraph{Oracle Query vs Single-hop Query}
We evaluate the oracle query against the single-hop query, i.e., querying with the original question, on the \hotpotqa{} dev set.
Specifically, we compare the recall of gold paragraphs, because the greater the recall, the fewer documents we need to pass into the expensive neural multi-hop QA component.

We index the English Wikipedia dump with introductory paragraphs provided by the \hotpotqa{} authors\footnote{\url{https://hotpotqa.github.io/wiki-readme.html}} with Elasticsearch 6.7 \cite{gormley2015elasticsearch},
where we index the titles and document text in separate fields with bigram indexing enabled.
This results in an index with 5,233,329 total documents.
At retrieval time, we boost the scores of any search result whose title matches the search query better -- this results in a better recall for entities with common names (\eg, ``\emph{Armada}'' the novel).
For more details about how the IR engine is set up and the effect of score boosting, please refer to Appendix \ref{sec:elasticsearch}.

\begin{figure}
    \pgfplotstableread[row sep=\\,col sep=&]{
	k & recall \\
	1 & 68.16 \\
	2 & 77.31 \\
	5 & 84.01 \\
	10 & 87.85 \\
	20 & 91.15 \\
	50 & 93.59 \\
}\questionpone

\pgfplotstableread[row sep=\\,col sep=&]{
	k & recall \\
	1 & 0.00 \\
	2 & 16.48 \\
	5 & 30.13 \\
	10 & 36.91 \\
	20 & 43.08 \\
	50 & 49.17 \\
}\questionptwo

\pgfplotstableread[row sep=\\,col sep=&]{
	k & recall \\
	1 & 85.36 \\
	2 & 90.52 \\
	5 & 94.53 \\
	10 & 95.87 \\
	20 & 97.12 \\
	50 & 98.15 \\
}\oraclepone

\pgfplotstableread[row sep=\\,col sep=&]{
	k & recall \\
	1 & 75.07 \\
	2 & 79.41 \\
	5 & 85.28 \\
	10 & 88.33 \\
	20 & 90.57 \\
	50 & 93.00 \\
}\oracleptwo

\begin{tikzpicture}[every plot/.append style={thick},font=\small]
\begin{axis}[
xlabel={Number of Retrieved Documents},
xtick={1, 2, 5, 10, 20, 50},
xticklabels={1,2,5,10,20,50},
xmode=log,
ylabel={Dev Recall (\%)},
height=5.5cm,
ymin=0,
ymax=100,
legend style={at={(0.5,-0.27)}, anchor=north},
legend columns=2,
ymajorgrids=true,
grid style=dashed,
]
\addplot[dashed, color=blue, mark=o, mark options={solid}] plot table[x=k, y=recall]{\questionpone};
\addplot[dashed, color=orange, mark=x, mark options={solid}] plot table[x=k,y=recall]{\questionptwo};
\addplot[color=blue, mark=o] plot table[x=k, y=recall]{\oraclepone};
\addplot[color=orange, mark=x] plot table[x=k,y=recall]{\oracleptwo};
\legend{Single-hop $d_1$, Single-hop $d_2$, Oracle $d_1$, Oracle $d_2$}

\end{axis}
\end{tikzpicture}
    \caption{Recall comparison between single-hop queries and \golden{} Retriever oracle queries for both supporting paragraphs on the \hotpotqa{} dev set.
    Note that the oracle queries are much more effective than the original question (single-hop query) at retrieving target paragraphs in both hops.}
    \label{fig:question_vs_oracle}
\end{figure}
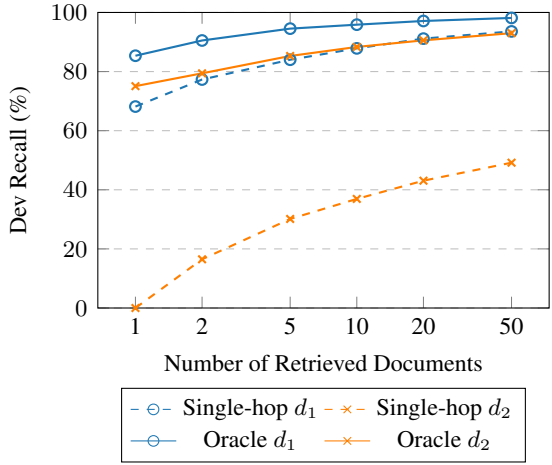

In Figure \ref{fig:question_vs_oracle}, we compare the recall of the two gold paragraphs required for each question in \hotpotqa{} at various number of documents retrieved (R@$n$) for the single-hop query and the multi-hop queries generated from the oracle.
Note that the oracle queries are much more effective at retrieving the gold paragraphs than the original question in both hops.
For instance, if we combine R@5 of both oracles (which effectively retrieves 10 documents from two queries) and compare that to R@10 for the single-hop query, the oracle queries improve recall for $d_1$ by 6.68\%, and that for $d_2$ by a significant margin of 49.09\%%
\footnote{Since \hotpotqa{} does not provide the logical order its gold entities should be discovered, we simply call the document $d_1$ which is more easily retrievable with the queries, and the other as $d_2$.}
This means that the final QA model will need to consider far fewer documents to arrive at a decent set of supporting facts that lead to the answer.

\begin{figure}
    \centering
    \includegraphics[width=0.48\textwidth]{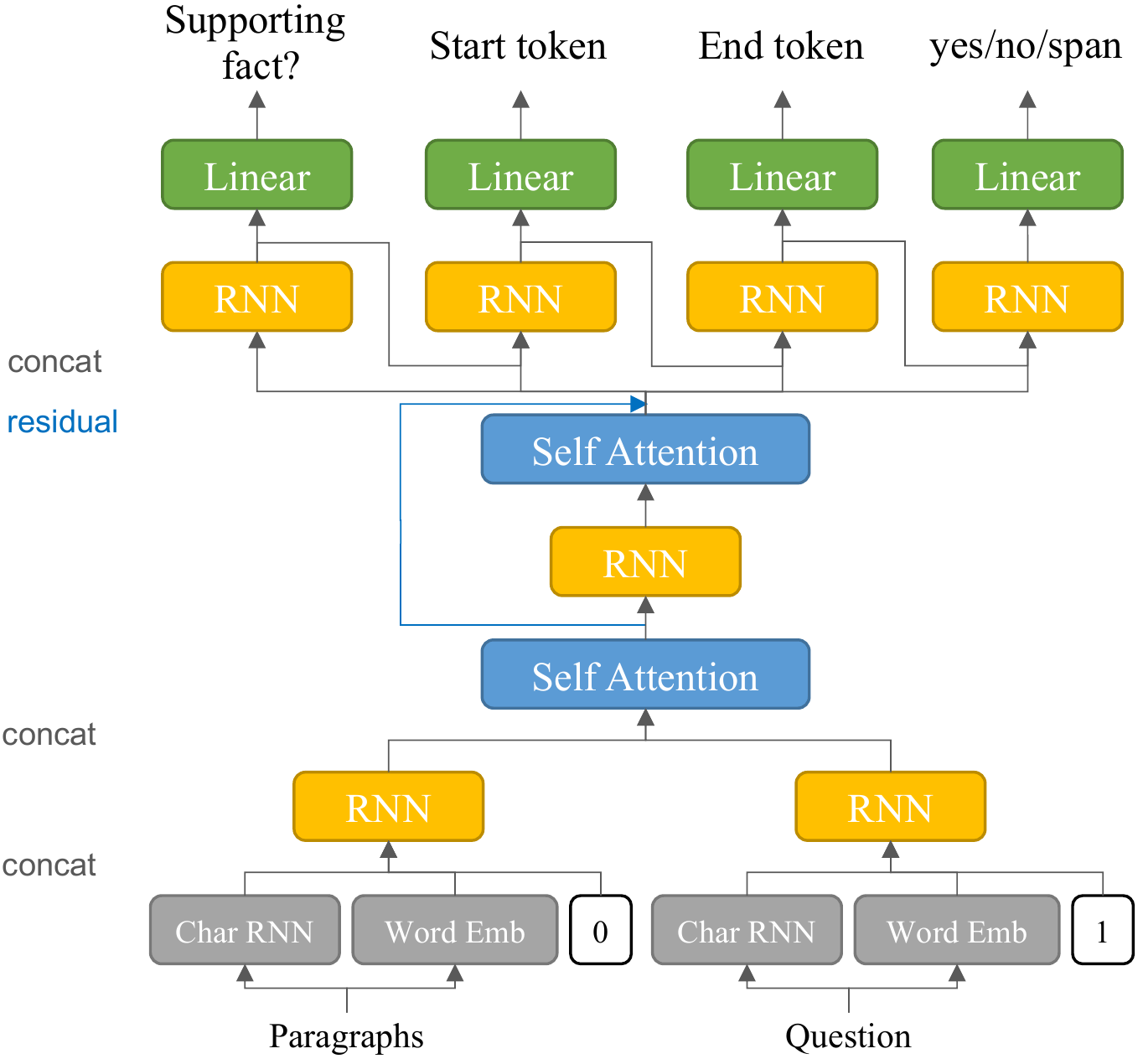}
    \caption{Question answering component in \golden{} Retriever. (Best viewed in color)} \label{fig:qa_model}
\end{figure}

\subsection{Question Answering Component} \label{sec:qa-component}

The final QA component of \golden{} Retriever is based on the baseline model presented in \cite{yang2018hotpotqa}, which is in turn based on BiDAF++ \cite{clark2017simple}.
We make two major changes to this model.
\citet{yang2018hotpotqa} concatenated all context paragraphs into one long string to predict span begin and end offsets for the answer, which is potentially sensitive to the order in which these paragraphs are presented to the model.
We instead process them separately with shared encoder RNN parameters to obtain paragraph order-insensitive representations for each paragraph.
Span offset scores are predicted from each paragraph independently before finally aggregated and normalized with a global softmax operation to produce probabilities over spans.
The second change is that we replace all attention mechanism in the original model with self attention layers over the concatenated question and context.
To differentiate context paragraph representations from question representations in this self-attention mechanism, we indicate question and context tokens by concatenating a 0/1 feature at the input layer.
Figure \ref{fig:qa_model} illustrates the QA model architecture.

\begin{table*}
    \centering
    \begin{tabular}{l ccc ccc }
    \toprule
    \multirow{2}{*}{\textbf{System}} & \multicolumn{2}{c}{\textbf{Answer}} & \multicolumn{2}{c}{\textbf{Sup Fact}} & \multicolumn{2}{c}{\textbf{Joint}} \\
    \cline{2-7}
    &EM&\fone{}&EM&\fone{}&EM& \fone{} \\
    \midrule
        Baseline \cite{yang2018hotpotqa}
            & 25.23 & 34.40 & \phantom{0}5.07 & 40.69 & \phantom{0}2.63 & 17.85 \\
        GRN + BERT & 29.87 & 39.14 & 13.16 & 49.67 & \phantom{0}8.26 & 25.84 \\
        MUPPET \cite{feldman2019multi} & 30.61 & 40.26 & 16.65 & 47.33 & 10.85 & 27.01 \\
        CogQA \cite{ding2019cognitive}
         &  37.12 & 48.87 & 22.82 & 57.69 & 12.42 & 34.92 \\
        PR-Bert & 43.33 & 53.79 & 21.90 & 59.63 & 14.50 & 39.11 \\
        Entity-centric BERT Pipeline & 41.82 & 53.09 & 26.26 & 57.29 & 17.01 & 39.18 \\
        BERT pip. (contemporaneous)
         & 45.32 & 57.34 & 38.67 & 70.83 & 25.14 & 47.60 \\
    \midrule
        {\golden{} Retriever}
        & \fake{37.92} & \fake{48.58} & \fake{30.69} & \fake{64.24} & \fake{18.04} & \fake{39.13} \\
    \bottomrule
    \end{tabular}
    \vspace{1mm}
    \caption{End-to-end QA performance of baselines and our \golden{} Retriever model on the \hotpotqa{} fullwiki test set.
    Among systems that were not published at the time of submission of this paper, ``BERT pip.'' was submitted to the official \hotpotqa{} leaderboard on May 15\textsuperscript{th} (thus contemporaneous), while ``Entity-centric BERT Pipeline'' and ``PR-Bert'' were submitted after the paper submission deadline.
    }
    \label{tab:end-to-end-perf}
\end{table*}

\section{Experiments}

\subsection{Data and Setup}

We evaluate our models in the fullwiki setting of \hotpotqa{} \cite{yang2018hotpotqa}.
\hotpotqa{} is a question answering dataset collected on the English Wikipedia, containing about 113k crowd-sourced questions that are constructed to require the introduction paragraphs of two Wikipedia articles to answer.
Each question in the dataset comes with the two gold paragraphs, as well as a list of sentences in these paragraphs that crowdworkers identify as supporting facts necessary to answer the question.
A diverse range of reasoning strategies are featured in \hotpotqa{}, including questions involving missing entities in the question (our \emph{Armada} example), intersection questions (\emph{What satisfies property A and property B?}), and comparison questions, where two entities are compared by a common attribute, among others.
In the few-document \emph{distractor} setting, the QA models are given ten paragraphs in which the gold paragraphs are guaranteed to be found; in the open-domain \emph{fullwiki} setting, which we focus on, the models are only given the question and the entire Wikipedia.
Models are evaluated on their answer accuracy and explainability, where the former is measured as overlap between the predicted and gold answers with exact match (EM) and unigram \fone{}, and the latter concerns how well the predicted supporting fact sentences match human annotation (Supporting Fact EM/\fone{}).
A joint metric is also reported on this dataset, which encourages systems to perform well on both tasks simultaneously.

We use the Stanford CoreNLP toolkit \cite{manning-EtAl:2014:P14-5} to preprocess Wikipedia, as well as to generate POS/NER features for the query generators, following \cite{yang2018hotpotqa} and \cite{chen2017reading}.
We always detokenize a generated search query before sending it to Elasticsearch, which has its own preprocessing pipeline.
Since all questions in \hotpotqa{} require exactly two supporting documents, we fix the number of retrieval steps of \golden{} Retriever to $S=2$.
To accommodate arbitrary steps of reasoning in \golden{} Retriever, a stopping criterion is required to determine when to stop retrieving for more supporting documents and perform few-document question answering, which we leave to future work.
During training and evaluation, we set the number of retrieved documents added to the retrieval context to 5 for each retrieval step, so that the total number of paragraphs our final QA model considers is 10, for a fair comparison to \cite{yang2018hotpotqa}.

We include hyperparameters and training details in Appendix \ref{sec:training} for reproducibility.

\subsection{End-to-end Question Answering}

We compare the end-to-end performance of \golden{} Retriever against several QA systems on the \hotpotqa{} dataset: (1) the baseline presented in \cite{yang2018hotpotqa}, (2) CogQA \cite{ding2019cognitive}, the top-performing previously published system, and (3) other high-ranking systems on the leaderboard.
As shown in Table \ref{tab:end-to-end-perf}, \golden{} Retriever is much better at locating the correct supporting facts from Wikipedia compared to CogQA, as well as most of the top-ranking systems. However, the QA performance is handicapped because we do not make use of pretrained contextualization models (\eg, BERT) that these systems use.
We expect a boost in QA performance from adopting these more powerful question answering models, especially ones that are tailored to perform few-document multi-hop reasoning. We leave this to future work.

\begin{table}[!t]
    \centering
    \small
    \begin{tabular}{lccc}
        \toprule
        \multicolumn{1}{c}{\textbf{Setting}} & \textbf{Ans \fone{}} & \textbf{Sup \fone{}} & \textbf{R@10$^*$} \\
        \midrule
        \golden{} Retriever & \fake{49.79} & \fake{64.58} & \fake{75.46} \\
        Single-hop query & \fake{38.19} & \fake{54.82} & 62.38 \\
        \hotpotqa{} IR & \fake{36.34} & \fake{46.78} & 55.71 \\
        \bottomrule
    \end{tabular}
    \caption{Question answering and IR performance amongst different IR settings on the dev set. We observe that although improving the IR engine is helpful, most of the performance gain results from the iterative retrieve-and-read strategy of \golden{} Retriever. (*: for \golden{} Retriever, the 10 paragraphs are combined from both hops, 5 from each hop.)} \label{tab:ir_ablation}
\end{table}

To understand the contribution of \golden{} Retriever's iterative retrieval process compared to that of the IR engine, we compare the performance of \golden{} Retriever against two baseline systems on the dev set: one that retrieves 10 supporting paragraphs from Elasticsearch with the original question, and one that uses the IR engine presented in \hotpotqa{}%
\footnote{For the latter, we use the fullwiki test input file the authors released, which contains the top-10 IR output from that retrieval system with the question as the query.}
In all cases, we use the QA component in \golden{} Retriever for the final question answering step.
As shown in Table \ref{tab:ir_ablation}, replacing the hand-engineered IR engine in \cite{yang2018hotpotqa} with Elasticsearch does result in some  gain in recall of the gold documents, but that does not translate to a significant improvement in QA performance.
Further inspection reveals that despite Elasticsearch improving overall recall of gold documents, it is only able to retrieve both gold documents for 36.91\% of the dev set questions, in comparison to 28.21\% from the IR engine in \cite{yang2018hotpotqa}.
In contrast, \golden{} Retriever improves this percentage to \fake{61.01\%},
almost doubling the recall over the single-hop baseline, providing the QA component a much better set of context documents to predict answers from.

\begin{table}
    \centering
    \small
    \begin{tabular}{lccc}
        \toprule
        \multicolumn{1}{c}{\textbf{System}} & \textbf{Ans \fone{}} & \textbf{Sup \fone{}} & \textbf{Joint \fone{}} \\
        \midrule
        \golden{} Retriever & \fake{49.79} & \fake{64.58} & \fake{40.21} \\
        w/ Hop 1 oracle & \fake{52.53} & \fake{68.06} & \fake{42.68} \\
        w/ Hop 1 \& 2 oracles & \fake{62.32} & \fake{77.00}  & \fake{52.18} \\
        \bottomrule
    \end{tabular}
    \caption{Pipeline ablative analysis of \golden{} Retriever end-to-end QA performance by replacing each query generator with a query oracle.} \label{tab:ablative}
\end{table}

Lastly, we perform an ablation study in which we replace our query generator models with our query oracles and observe the effect on end-to-end performance.
As can be seen in Table \ref{tab:ablative}, replacing $G_1$ with the oracle only slightly improves end-to-end performance, but further substituting $G_2$ with the oracle yields a significant improvement.
This illustrates that the performance loss is largely attributed to $G_2$ rather than $G_1$, because $G_2$ solves a harder span selection problem from a longer retrieval context.
In the next section, we examine the query generation models more closely by evaluating their performance without the QA component.

\subsection{Analysis of Query Generation}

\begin{table*}
    \centering
    \resizebox{1\textwidth}{!}{
    \small
    \begin{tabular}{cp{7cm}p{4.15cm}p{3.4cm}}
        \toprule
        &
        \multicolumn{1}{c}{\textbf{Question}} & \textbf{Predicted $q_1$} & \textbf{Predicted $q_2$} \\
        \midrule
        (1) & What video game character did the voice actress in the animated film Alpha and Omega voice? &	voice actress in the animated film Alpha and Omega \emph{\color{blue}(animated film Alpha and Omega voice)} &	Hayden Panettiere \\
        \midrule
        (2) & What song was created by the group consisting of Jeffrey Jey, Maurizio Lobina and Gabry Ponte and released on 15 January 1999?& Jeffrey Jey \emph{\color{blue}(group consisting of Jeffrey Jey, Maurizio Lobina and Gabry Ponte)} &	Gabry Ponte and released on 15 January 1999 \emph{\color{blue}(``Blue (Da Ba Dee)'')} \\
        \midrule
        (3) & Yau Ma Tei North is a district of a city with how many citizens? & Yau Ma Tei North & Yau Tsim Mong District of Hong Kong \emph{\color{blue}(Hong Kong)} \\
        \midrule
        (4) & What company started the urban complex development that included the highrise building, The Harmon? & highrise building, The Harmon & CityCenter\\
        \bottomrule
    \end{tabular}
    }
    \caption{Examples of predicted queries from the query generators on the \hotpotqa{} dev set. The oracle query is displayed in blue in parentheses if it differs from the predicted one.}
    \label{tab:predicted_queries}
\end{table*}

To evaluate the query generators, we begin by determining how well they emulate the oracles.
We evaluate them using Exact Match (EM) and \fone{} on the span prediction task, as well as compare their queries' retrieval performance against the oracle queries.
As can be seen in Table \ref{tab:query_generator_perf}, the performance of $G_2$ is worse than that of $G_1$ in general, confirming our findings on the end-to-end pipeline.
When we combine them into a pipeline, the generated queries perform only slightly better on $d_1$ when a total of 10 documents are retrieved (\fake{89.91\%} vs 87.85\%), but are significantly more effective for $d_2$ (\fake{61.01\%} vs 36.91\%).
If we further zoom in on the retrieval performance on non-comparison questions for which finding the two entities involved is less trivial, we can see that the recall on $d_2$ improves from 27.88\% to \fake{53.23\%}, almost doubling the number of questions we have the complete gold context to answer.
We note that the IR performance we report on the full pipeline is different to that when we evaluate the query generators separately. We attribute this difference to the fact that the generated queries sometimes retrieve both gold documents in one step.

\begin{table}
    \centering
    \begin{tabular}{cccc}
    \toprule
    \multicolumn{1}{c}{\multirow{2}{*}{\textbf{Model}}} & \multicolumn{2}{c}{\textbf{Span}} & \multicolumn{1}{c}{\multirow{2}{*}{\textbf{R@5}}}\\
    \cline{2-3}
         & EM & \fone{} & \\
        \midrule
        $G_1$ & 51.40 & 78.75 & 85.86 \\
        $G_2$ & \fake{52.29} & \fake{63.07} & \fake{64.83} \\
        \bottomrule
    \end{tabular}
    \caption{Span prediction and IR performance of the query generator models for Hop 1 ($G_1$) and Hop 2 ($G_2$) evaluated separately on the \hotpotqa{} dev set.} \label{tab:query_generator_perf}
\end{table}

To better understand model behavior, we also randomly sampled some examples from the dev set to compare the oracle queries and the predicted queries.
Aside from exact matches, we find that the predicted queries are usually small variations of the oracle ones.
In some cases, the model selects spans that are more natural and informative (Example (1) in Table \ref{tab:predicted_queries}).
When they differ a bit more, the model is usually overly biased towards shorter entity spans and misses out on informative information (Example (2)).
When there are multiple entities in the retrieval context, the model sometimes selects the wrong entity, which suggests that a more powerful query generator might be desirable (Example (3)).
Despite these issues, we find that these natural language queries make the reasoning process more interpretable, and easier for a human to verify or intervene as needed.

\paragraph{Limitations}
Although we have demonstrated that generating search queries with span selection works in most cases, it also limits the kinds of queries we can generate, and in some cases leads to undesired behavior.
One common issue is that the entity of interest has a name shared by too many Wikipedia pages (\eg, ``\emph{House Rules}'' the 2003 TV series).
This sometimes results in the inclusion of extra terms in our oracle query to expand it (\eg, Example (4) specifies that ``\emph{The Harmon}'' is a highrise building).
In other cases, our span oracle makes use of too much information from the gold entities (Example (2), where a human would likely query for ``\emph{Eiffel 65 song released 15 January 1999}'' because ``\emph{Blue}'' is not the only song mentioned in $d_1$).
We argue, though, these are due to the simplifying choice of span selection for query generation and fixed number of query steps. We leave extensions to future work.

\section{Conclusion}

In this paper, we presented \golden{} (Gold Entity) Retriever, an open-domain multi-hop question answering system for scalable multi-hop reasoning.
Through iterative reasoning and retrieval, \golden{} Retriever greatly improves the recall of gold supporting facts, thus providing the question answering model a much better set of context documents to produce an answer from, and demonstrates competitive performance to the state of the art.
Generating natural languages queries for each step of reasoning, \golden{} Retriever is also more interpretable to humans compared to previous neural retrieval approaches and affords better understanding and verification of model behavior.

\section*{Acknowledgements}

The authors would like to thank Robin Jia among other members of the Stanford NLP Group, as well as the anonymous reviewers for discussions and comments on earlier versions of this paper.
Peng Qi would also like to thank Suprita Shankar, Jamil Dhanani, and Suma Kasa for early experiments on BiDAF++ variants for \hotpotqa{}.
This research is funded in part by Samsung Electronics Co., Ltd.\ and in part by the SAIL-JD Research Initiative.

\bibliography{deep-retriever}

\begin{thebibliography}{30}
\expandafter\ifx\csname natexlab\endcsname\relax\def\natexlab#1{#1}\fi

\bibitem[{Buck et~al.(2018)Buck, Bulian, Ciaramita, Gajewski, Gesmundo,
  Houlsby, and Wang}]{buck2017ask}
Christian Buck, Jannis Bulian, Massimiliano Ciaramita, Wojciech Gajewski,
  Andrea Gesmundo, Neil Houlsby, and Wei Wang. 2018.
\newblock Ask the right questions: Active question reformulation with
  reinforcement learning.
\newblock In \emph{Proceedings of the International Conference on Learning
  Representations}.

\bibitem[{Chen et~al.(2017)Chen, Fisch, Weston, and Bordes}]{chen2017reading}
Danqi Chen, Adam Fisch, Jason Weston, and Antoine Bordes. 2017.
\newblock Reading {Wikipedia} to answer open-domain questions.
\newblock In \emph{Association for Computational Linguistics (ACL)}.

\bibitem[{Chen and Durrett(2019)}]{chen2019understanding}
Jifan Chen and Greg Durrett. 2019.
\newblock Understanding dataset design choices for multi-hop reasoning.
\newblock In \emph{Proceedings of the Conference of the North American Chapter
  of the Association for Computational Linguistics ({NAACL})}.

\bibitem[{Clark and Gardner(2017)}]{clark2017simple}
Christopher Clark and Matt Gardner. 2017.
\newblock Simple and effective multi-paragraph reading comprehension.
\newblock In \emph{Proceedings of the 55th Annual Meeting of the Association of
  Computational Linguistics}.

\bibitem[{Das et~al.(2019)Das, Dhuliawala, Zaheer, and
  McCallum}]{das2019multistep}
Rajarshi Das, Shehzaad Dhuliawala, Manzil Zaheer, and Andrew McCallum. 2019.
\newblock \href {https://openreview.net/forum?id=HkfPSh05K7} {Multi-step
  retriever-reader interaction for scalable open-domain question answering}.
\newblock In \emph{International Conference on Learning Representations}.

\bibitem[{De~Cao et~al.(2019)De~Cao, Aziz, and Titov}]{decao2019question}
Nicola De~Cao, Wilker Aziz, and Ivan Titov. 2019.
\newblock Question answering by reasoning across documents with graph
  convolutional networks.
\newblock In \emph{Proceedings of the Conference of the North American Chapter
  of the Association for Computational Linguistics ({NAACL})}.

\bibitem[{Devlin et~al.(2019)Devlin, Chang, Lee, and
  Toutanova}]{devlin2018bert}
Jacob Devlin, Ming-Wei Chang, Kenton Lee, and Kristina Toutanova. 2019.
\newblock {BERT}: Pre-training of deep bidirectional transformers for language
  understanding.
\newblock In \emph{Proceedings of the Conference of the North American Chapter
  of the Association for Computational Linguistics ({NAACL})}.

\bibitem[{Ding et~al.(2019)Ding, Zhou, Chen, Yang, and
  Tang}]{ding2019cognitive}
Ming Ding, Chang Zhou, Qibin Chen, Hongxia Yang, and Jie Tang. 2019.
\newblock Cognitive graph for multi-hop reading comprehension at scale.
\newblock In \emph{Proceedings of the 57th Annual Meeting of the Association of
  Computational Linguistics}.

\bibitem[{Feldman and El-Yaniv(2019)}]{feldman2019multi}
Yair Feldman and Ran El-Yaniv. 2019.
\newblock Multi-hop paragraph retrieval for open-domain question answering.
\newblock In \emph{Proceedings of the 57th Annual Meeting of the Association of
  Computational Linguistics}.

\bibitem[{Gormley and Tong(2015)}]{gormley2015elasticsearch}
Clinton Gormley and Zachary Tong. 2015.
\newblock \emph{Elasticsearch: The definitive guide: A distributed real-time
  search and analytics engine}.
\newblock O'Reilly Media, Inc.

\bibitem[{Hassan~Awadallah et~al.(2014)Hassan~Awadallah, White, Pantel, Dumais,
  and Wang}]{hassan2014supporting}
Ahmed Hassan~Awadallah, Ryen~W White, Patrick Pantel, Susan~T Dumais, and
  Yi-Min Wang. 2014.
\newblock Supporting complex search tasks.
\newblock In \emph{Proceedings of the 23rd ACM International Conference on
  Conference on Information and Knowledge Management}, pages 829--838. ACM.

\bibitem[{Joshi et~al.(2017)Joshi, Choi, Weld, and
  Zettlemoyer}]{JoshiTriviaQA2017}
Mandar Joshi, Eunsol Choi, Daniel~S. Weld, and Luke Zettlemoyer. 2017.
\newblock {TriviaQA}: A large scale distantly supervised challenge dataset for
  reading comprehension.
\newblock In \emph{Proceedings of the 55th Annual Meeting of the Association
  for Computational Linguistics}.

\bibitem[{Kingma and Ba(2015)}]{kingma2015adam}
Diederik~P. Kingma and Jimmy Ba. 2015.
\newblock Adam: A method for stochastic optimization.
\newblock In \emph{Proceedings of the International Conference on Learning
  Representations ({ICLR})}.

\bibitem[{Kratzwald and Feuerriegel(2018)}]{Kratzwald2018AdaptiveDR}
Bernhard Kratzwald and Stefan Feuerriegel. 2018.
\newblock Adaptive document retrieval for deep question answering.
\newblock In \emph{Proceedings of the Conference on Empirical Methods in
  Natural Language Processing ({EMNLP})}.

\bibitem[{Manning et~al.(2014)Manning, Surdeanu, Bauer, Finkel, Bethard, and
  McClosky}]{manning-EtAl:2014:P14-5}
Christopher~D. Manning, Mihai Surdeanu, John Bauer, Jenny Finkel, Steven~J.
  Bethard, and David McClosky. 2014.
\newblock \href {http://www.aclweb.org/anthology/P/P14/P14-5010} {The
  {Stanford} {CoreNLP} natural language processing toolkit}.
\newblock In \emph{Association for Computational Linguistics (ACL) System
  Demonstrations}, pages 55--60.

\bibitem[{Mehrotra et~al.(2016)Mehrotra, Bhattacharya, and
  Yilmaz}]{mehrotra2016deconstructing}
Rishabh Mehrotra, Prasanta Bhattacharya, and Emine Yilmaz. 2016.
\newblock Deconstructing complex search tasks: A bayesian nonparametric
  approach for extracting sub-tasks.
\newblock In \emph{Proceedings of the 2016 Conference of the North American
  Chapter of the Association for Computational Linguistics: Human Language
  Technologies}, pages 599--605.

\bibitem[{Mehrotra and Yilmaz(2017)}]{mehrotra2017extracting}
Rishabh Mehrotra and Emine Yilmaz. 2017.
\newblock Extracting hierarchies of search tasks \& subtasks via a bayesian
  nonparametric approach.
\newblock In \emph{Proceedings of the 40th International ACM SIGIR Conference
  on Research and Development in Information Retrieval}, pages 285--294. ACM.

\bibitem[{Min et~al.(2019{\natexlab{a}})Min, Wallace, Singh, Gardner,
  Hajishirzi, and Zettlemoyer}]{min2019compositional}
Sewon Min, Eric Wallace, Sameer Singh, Matt Gardner, Hannaneh Hajishirzi, and
  Luke Zettlemoyer. 2019{\natexlab{a}}.
\newblock Compositional questions do not necessitate multi-hop reasoning.
\newblock In \emph{Proceedings of the Annual Conference of the Association of
  Computational Linguistics}.

\bibitem[{Min et~al.(2019{\natexlab{b}})Min, Zhong, Zettlemoyer, and
  Hajishirzi}]{min2019multi}
Sewon Min, Victor Zhong, Luke Zettlemoyer, and Hannaneh Hajishirzi.
  2019{\natexlab{b}}.
\newblock Multi-hop reading comprehension through question decomposition and
  rescoring.
\newblock In \emph{Proceedings of the Annual Conference of the Association of
  Computational Linguistics}.

\bibitem[{Nishida et~al.(2018)Nishida, Saito, Otsuka, Asano, and
  Tomita}]{nishida2018retrieve}
Kyosuke Nishida, Itsumi Saito, Atsushi Otsuka, Hisako Asano, and Junji Tomita.
  2018.
\newblock Retrieve-and-read: Multi-task learning of information retrieval and
  reading comprehension.
\newblock In \emph{Proceedings of the 27th ACM International Conference on
  Information and Knowledge Management}, pages 647--656. ACM.

\bibitem[{Nogueira and Cho(2017)}]{nogueira2017task}
Rodrigo Nogueira and Kyunghyun Cho. 2017.
\newblock Task-oriented query reformulation with reinforcement learning.

\bibitem[{Nogueira et~al.(2019)Nogueira, Yang, Lin, and
  Cho}]{nogueira2019document}
Rodrigo Nogueira, Wei Yang, Jimmy Lin, and Kyunghyun Cho. 2019.
\newblock Document expansion by query prediction.
\newblock \emph{arXiv preprint arXiv:1904.08375}.

\bibitem[{Rajpurkar et~al.(2018)Rajpurkar, Jia, and Liang}]{rajpurkar2018know}
Pranav Rajpurkar, Robin Jia, and Percy Liang. 2018.
\newblock Know what you don't know: Unanswerable questions for {SQuAD}.
\newblock In \emph{Proceedings of the 56th Annual Meeting of the Association
  for Computational Linguistics}.

\bibitem[{Rajpurkar et~al.(2016)Rajpurkar, Zhang, Lopyrev, and
  Liang}]{rajpurkar2016squad}
Pranav Rajpurkar, Jian Zhang, Konstantin Lopyrev, and Percy Liang. 2016.
\newblock {SQuAD}: 100,000+ questions for machine comprehension of text.
\newblock In \emph{Proceedings of the Conference on Empirical Methods in
  Natural Language Processing ({EMNLP})}.

\bibitem[{Robertson et~al.(1995)Robertson, Walker, Jones, Hancock-Beaulieu,
  Gatford et~al.}]{robertson1995okapi}
Stephen~E. Robertson, Steve Walker, Susan Jones, Micheline~M Hancock-Beaulieu,
  Mike Gatford, et~al. 1995.
\newblock Okapi at trec-3.
\newblock \emph{Nist Special Publication Sp}, 109:109.

\bibitem[{Sutton et~al.(2000)Sutton, McAllester, Singh, and
  Mansour}]{sutton2000policy}
Richard~S. Sutton, David~A. McAllester, Satinder~P. Singh, and Yishay Mansour.
  2000.
\newblock Policy gradient methods for reinforcement learning with function
  approximation.
\newblock In \emph{Advances in Neural Information Processing Systems}, pages
  1057--1063.

\bibitem[{Talmor and Berant(2018)}]{talmor2018web}
Alon Talmor and Jonathan Berant. 2018.
\newblock The web as a knowledge-base for answering complex questions.
\newblock \emph{Proceedings of the Conference of the North American Chapter of
  the Association for Computational Linguistics ({NAACL})}.

\bibitem[{Welbl et~al.(2018)Welbl, Stenetorp, and
  Riedel}]{welbl2018constructing}
Johannes Welbl, Pontus Stenetorp, and Sebastian Riedel. 2018.
\newblock Constructing datasets for multi-hop reading comprehension across
  documents.
\newblock \emph{Transactions of the Association of Computational Linguistics},
  6:287--302.

\bibitem[{Yang et~al.(2018)Yang, Qi, Zhang, Bengio, Cohen, Salakhutdinov, and
  Manning}]{yang2018hotpotqa}
Zhilin Yang, Peng Qi, Saizheng Zhang, Yoshua Bengio, William~W. Cohen, Ruslan
  Salakhutdinov, and Christopher~D. Manning. 2018.
\newblock {HotpotQA}: A dataset for diverse, explainable multi-hop question
  answering.
\newblock In \emph{Proceedings of the Conference on Empirical Methods in
  Natural Language Processing ({EMNLP})}.

\bibitem[{Zhong et~al.(2019)Zhong, Xiong, Keskar, and Socher}]{zhong2019coarse}
Victor Zhong, Caiming Xiong, Nitish~Shirish Keskar, and Richard Socher. 2019.
\newblock Coarse-grain fine-grain coattention network for multi-evidence
  question answering.
\newblock In \emph{Proceedings of the International Conference on Learning
  Representations}.

\end{thebibliography}
\bibliographystyle{acl_natbib}

\clearpage
\appendix
\section{Elasticsearch Setup} \label{sec:elasticsearch}

\subsection{Setting Up the Index}

We start from the Wikipedia dump file containing the introductory paragraphs used in \hotpotqa{} that \citet{yang2018hotpotqa} provide,\footnote{\url{https://hotpotqa.github.io/wiki-readme.html}} and add the fields corresponding to Wikipedia page titles and the introductory paragraphs (text) into the index.

For the title, we use Elasticsearch's \texttt{simple} analyzer which performs basic tokenization and lowercasing of the content.
For the text, we join all the sentences and use the \texttt{standard} analyzer which further allows for removal of punctuation and stop words.
For both fields, we index an auxiliary field with bigrams using the \texttt{shingle} filter,\footnote{\url{https://www.elastic.co/guide/en/elasticsearch/reference/6.7/analysis-shingle-tokenfilter.html}} and perform basic \texttt{asciifolding} to map non ASCII characters to a similar ASCII character (\eg, ``\'{e}''$\to$ ``e'').

At search time, we launch a \texttt{multi\_match} query against all fields with the same query, which performs a full-text query employing the BM25 ranking function \cite{robertson1995okapi} with all fields in the index, and returns the score of the best field for ranking by default.
To promote documents whose title match the search query, we boost the search score of all title-related fields by 1.25 in this query.

\subsection{Reranking Search Results}

In Wikipedia, it is common that pages or entity names share rare words that are important to search engines, and a naive full-text search IR system will not be able to pick the one that matches the query the best.
For instance, if one set up Elasticsearch according to the instructions above and searched for ``\emph{George W. Bush}'', he/she would be surprised to see that the actual page is not even in the top-10 search results, which contains entities such as ``\emph{George W. Bush Childhood Home}'' and ``\emph{Bibliography of George W. Bush}''.

To this end, we propose to rerank these query results with a simple but effective heuristic that alleviates this issue.
We would first retrieve at least 50 candidate documents for each query for consideration, and boost the query scores of documents whose title exactly matches the search query, or is a substring of the search query.
Specifically, we multiply the document score by a heuristic constant between 1.05 and 1.5, depending on how well the document title matches the query, before reranking all search results.
This results in a significant improvement in these cases.
For the query ``\emph{George W. Bush}'', the page for the former US president is ranked at the top after reranking.
In Table \ref{tab:ir_setup}, we also provide results from the single-hop query to show the improvement from title score boosting introduced from the previous section and reranking.

\begin{table}
    \centering
    \small
    \begin{tabular}{lcc}
        \toprule
        \multicolumn{1}{c}{\textbf{IR System}} & \textbf{R@10 for $d_1$} & \textbf{R@10 for $d_2$} \\
        \midrule
        Final system & 87.85 & 36.91 \\
        w/o Title Boosting & 86.85 & 32.64 \\
        w/o Reranking & 86.32 & 34.77 \\
        w/o Both & 84.67 & 29.55 \\
        \bottomrule
    \end{tabular}
    \caption{IR performance (recall in percentages) of various Elasticsearch setups on the \hotpotqa{} dev set using the original question.} \label{tab:ir_setup}
\end{table}

\section{Oracle Query Generation} \label{sec:heuristics}

We mainly employ three heuristics to find the semantic overlap between the retrieval context and the desired documents: longest common subsequence (LCS), longest common substring (LCSubStr), and overlap merging which generalizes the two.
Specifically, the overlap merging heuristic looks for contiguous spans in the retrieval context that have high rates of overlapping tokens with the desired document, determined by the total number of overlapping tokens divided by the total number of tokens considered in the span.

In all heuristics, we ignore stop words and lowercase the rest in computing the spans to capture more meaningful overlaps, and finally take the span in the retrieval context that all the overlapping words are contained in.
For instance, if the retrieval context contains ``\emph{the \golden{} Retriever model on \hotpotqa{}}'' and the desired document contains ``\emph{\golden{} Retriever on the \hotpotqa{} dataset}'', we will identify the overlapping terms as ``\golden{}'', ``\emph{Retriever}'', and ``\hotpotqa{}'', and return the span ``\emph{\golden{} Retriever model on \hotpotqa{}}'' as the resulting candidate query.

To generate candidates for the oracle query, we apply the heuristics between combinations of \{cleaned question, cleaned question without punctuation\} $\times$ \{cleaned document title, cleaned paragraph\}, where cleaning means stop word removal and lowercasing.
Once oracle queries are generated, we launch these queries against Elasticsearch to determine the rank of the desired paragraph.
If multiple candidate queries are able to place the desired paragraph in the top 5 results, we further rank the candidate queries by other metrics (\eg, length of the query) to arrive at the final oracle query to train the query generators.
We refer the reader to our released code for further details.

\section{Training Details} \label{sec:training}

\subsection{Query Generators}

Once the oracle queries are generated, we train our query generators to emulate them on the training set, and choose the best model with \fone{} in span selection on the dev set.
We experiment with hyperparameters such as learning rate, training epochs, batch size, number of word embeddings to finetune, among others, and report the final hyperparameters for both query generators ($G_1$ and $G_2$) in Table \ref{tab:query_generator_hyperparam}.

\begin{table}
\small
\centering
    \begin{tabular}{ll}
        \toprule

    Hyperparameter & Values \\
    \midrule
    Learning rate & \underline{$\mathit{5\times 10^{-4}}$}, $\mathbf{1\times10^{-3}}$ \\
    Finetune embeddings &  0, 200, \textbf{500}, \textit{\underline{1000}} \\
    Epoch & \textbf{25}, \textit{\underline{40}} \\
    Batch size &  \textit{\underline{\textbf{32}}}, 64, 128 \\
    Hidden size &  64, \textit{\underline{\textbf{128}}}, 256, 512, 768 \\
    Max sequence length &  15, \textit{\underline{20}}, 30, \textbf{50}, 100 \\
    Dropout rate &  0.3, 0.35, \textit{\underline{\textbf{0.4}}}, 0.45\\
    \bottomrule
    \end{tabular}
    \caption{Hyperparameter settings for the query generators. The final hyperparameters for the Hop 1 query generator are shown in \textbf{bold}, and those for the Hop 2 query generator are shown in \textit{\underline{underlined itallic}}.}
    \label{tab:query_generator_hyperparam}
\end{table}

\subsection{Question Answering Model}

Our final question answering component is trained with the paragraphs produced by the oracle queries (5 from each hop, 10 in total), with $d_1$ and $d_2$ inserted to replace the lowest ranking paragraph in each hop if they are not in the set already.

We develop our model based on the baseline model of \citet{yang2018hotpotqa}, and reuse the same default hyperparameters whenever possible.
The main differences in the hyperparameters are: we optimize our QA model with Adam (with default hyperparameters) \cite{kingma2015adam} instead of stochastic gradient descent with a larger batch size of 64; we anneal the learning rate by 0.5 with a patience of 3 instead of 1, that is, we multiply the learning rate by 0.5 after three consecutive failures to improve dev \fone{}; we clip the gradient down to a maximum $\ell_2$ norm of 5; we apply a 10\% dropout to the model, for which we have increased the hidden size to 128; and use 10 as the coefficient by which we multiply the supporting facts loss, before mixing it with the span prediction loss.
We configure the model to read 10 context paragraphs, and limit each paragraph to at most 400 tokens including the title.

\end{document}